\documentclass{ifacconf}
\usepackage{booktabs}
\usepackage{graphicx}      
\usepackage{natbib}        
\usepackage{multirow}
\usepackage{amsmath}
\usepackage{amsfonts}
\begin{document}
\begin{frontmatter}
\title{\small This work has been submitted to IFAC for possible publication}
\title{Water Supply Prediction Based on Initialized Attention Residual Network}

\thanks[footnoteinfo]{This work is supported by National Natural Science Foundation of China (No.61533013, 61633019, 61433002), Key projects from Ministry of Science and Technology (No.2017ZX07207003, 2017ZX07207005), Shaanxi Provincial Key Project (2018ZDXM-GY-168) and Shanghai Project (17DZ1202704).}
\author[First]{Yuhao Long}
\author[First,Second]{Jingcheng Wang}
\author[First]{Jingyi Wang}
\address[First]{Automation Department, Shanghai Jiao Tong University, Key Laboratory of System Control and Information Processing, Ministry of Education of China, Shanghai, 200240, China. (e-mail: Yuhao Long, 18608098621@163.com; Jingcheng Wang, jcwang@sjtu.edu.cn; Jingyi Wang, jy\_wang@sjtu.edu.cn)}
\address[Second]{Autonomous Systems and Intelligent Control International Joint Research Center, Xi'an Technological University, Xi'an, Shaanxi, 710021, China (e-mail: jcwang@sjtu.edu.cn)}

\begin{abstract}                
Real-time and accurate water supply forecast is crucial for water plant. However, most existing methods are likely affected by factors such as weather and holidays, which lead to a decline in the reliability of water supply prediction. In this paper, we address a generic artificial neural network, called Initialized Attention Residual Network (IARN), which is combined with an attention module and residual modules. Specifically, instead of continuing to use the recurrent neural network (RNN) in time-series tasks, we try to build a convolution neural network (CNN) to recede the disturb from other factors, relieve the limitation of memory size and get a more credible results. Our method achieves state-of-the-art performance on several datasets, in terms of accuracy, robustness and generalization ability.
\end{abstract}


\end{frontmatter}

\section{Introduction}\label{section1}
Water plays an important role in human life. How to model and predict the water consumption accurately are quite complicated and also very important. In practice, the daily and hourly forecasting is more widely needed. On the one hand, in order to help water plant carry out fast and efficient dispatch control, accurate hourly water forecasts are needed. On the other hand, we also need to help the water plant to carry out future water supply dispatch planning by providing reliable water forecast data. With the development of water supply prediction reaserch, water plant will be further intelligent and energy saving. Meanwhile, some abnormal conditions such as pipeline leakage can be warned earlier by precise water supply prediction proposed in the plant. Comparing the actual value with the predicted value, we can monitor the abnormal conditions promptly. Generally, water supply forecast is the premise and foundation of water supply system control and guidance, which is also one of the most important feature of the smart water management system.

Currently, support vector machine (SVM) (\cite{Ji2014Gravitational}; \cite{Wang2015Urban}), autoregressive moving average model (ARMA) and autoregressive integrated moving average model (ARIMA) (\cite{Colak2015Multi}) are commonly employed to forecast short-term water supply. These traditional methods are capable of achieving tolerable results on small-size datasets. However, there are still some aspects to be considered in these methods, such as low accuracy, poor robustness and generalization.
As shown in \cite{cross2016incremental} and \cite{Ullah2017Action}, some researchers tried to use deep learning structure based on recurrent neural network, for instance long short-term memory (LSTM) and bi-directional long short-term memory (Bi-LSTM) to get better results. Long-term dependencies are solved by LSTM with implementation of forget gate, input gate and output gate. Nevertheless, the data are steered from left to right side in LSTM, so the follow-up data present a higher weight than the previous. To solve this problem, backward LSTM is added to Bi-LSTM, generating the same weight to the front and rear information during the process. Nonetheless, some problems are still remained. First and foremost, both LSTM and GRU have already improved memory length of the basis of RNN but there is still limitation on their memory size. Secondly, these networks have requirements for the interrelationship among the data, such as the connection between the subject and the guest in the language, but the connection in dataset is not strong enough. Last but not the least, RNN for time-series tasks requires iterative training, which imports cumulative error step by step. While lots of extensive works have been done, recent temporal models have been limited to sliding window action detectors, see \cite{Ni2016Progressively}, \cite{Singh2016A} and \cite{Rohrbach2016Recognizing}, which typically do not capture long-range temporal patterns.

\begin{figure}[ht]
\begin{center}
\includegraphics[width=9.2cm]{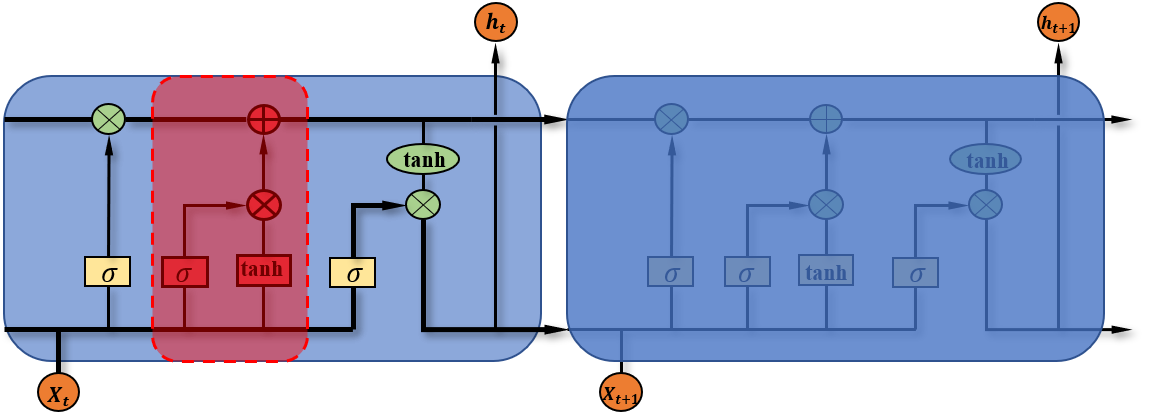}    
\caption{The structure of LSTM. Details are showed in different colors. $\oplus$ and $\otimes$ represent element-wise sum and multiplication operations, while $\sigma$ and $tanh$ are the optimizer layers in the network.} \label{fig_lstm}
\label{fig:bifurcation}
\end{center}
\end{figure}

 Informed by \cite{dauphin2017language}, \cite{kalchbrenner2016neural} and \cite{DBLP:journals/corr/OordDZSVGKSK16}, this paper introduces a novel artificial neural network based on CNN structure for overcoming these problems. Meanwhile, our model is also inspired by LSTM's structure as shown in Fig \ref{fig_lstm}. As we can see, the red rectangle frame is similar to the attention module which we are familiar with(see Fig \ref{fig_attention}), and the bold line is consistent with the residual  module(see Fig \ref{fig_res}). Unlike LSTM network, we model our network by directly combing the attention module with residual block so that our model can make full use of the existing information. To dispose the first problem, we employ the convolution-based model instead of RNN architecture. To handle the second issue, we apply a deep convolutional neural network, where the senior semantic information can be simply studied. And the residual block is adopted in our model to cope with the accumulated error. Thus, the processed network can completely meet the requirements as we mentioned above. Actually, the processed network has state-of-the-art results of water supply prediction. It is worth pointing that our network also has a good generalization ability and industrial application potential.

Our main contributions can be summarized as follows:
\begin{itemize}
    \item We devise a novel attention module which is suitable for flow prediction. Through the attention module, data preprocessing and data augmentation are equipped on the network. Therefore, important information will be emphasized while redundant information will be suppressed in training.
    \item We successfully apply the residual network (ResNet) in computer version (CV) to water supply prediction through transfer learning. And this allows our model to have the ability to take full use of the past information on processing. In addition, vanishing gradient and exploding gradient are avoided by the residual module.
        \item Our proposed method outperforms current state-of-the-art water supply prediction methods, even in small size of datasets. Meanwhile, our network has the advantages of outstanding generalization and robustness, which are appropriate for online operation.
\end{itemize}
This paper is organized as follows. In Section \ref{section2}, we mainly present the overall structure of our network and then introduced the details and functions of each module. The training and testing results of our network are presented in Section \ref{section3}. And Section \ref{section4} concludes this paper.

\section{Initialized ATTENTION RESIDUAL NETWORK}\label{section2}

Generally, RNN is commonly applied to time-series tasks. However, there is no such strong connection between the flow data and water supply data. The predictions are only based on the past information in the process of training. Therefore, we consider using convolution architecture to accomplish the prediction of water supply.

Suppose that we are given an input sequence $x_{0}, x_{1}, \ldots, x_{T}$ and wish to predict some corresponding outputs $y_{0}, \dots, y_{T}$ at each time.
The prediction $y_t$ is constrained only be related to those inputs $x_{0}, x_{1}, \ldots, x_{t}$. Time-series modeling network is any function $f: \mathcal{X}^{T+1} \rightarrow \mathcal{Y}^{T+1}$ that generate this mapping :
 $x_{0}, x_{1}, \ldots, x_{t}$:
\begin{equation}
\hat{y}_{0}, \ldots, \hat{y}_{T}=f\left(x_{0}, \ldots, x_{T}\right)
\end{equation}
In this function, there is no information `omission' from future to past. $y_{t}$ only depends on $x_{0}, x_{1}, \ldots, x_{t}$ without relying on any `future' inputs
$x_{t+1}, x_{t+2}, \ldots, x_{T}$. We aim to find a network function that minimizes some expected loss between the actual value and predictions.

Taking into account the convolution structure, we need to stack convolution layers of a certain depth to satisfy the requirements of feature extraction in our time-series data. However, this may lead to vanishing/exploding gradients which can be found in \cite{BENGIO1994Learning} and \cite{He2015Convolutional}. In the field of computer vision, the residual network structures (\cite{he2016deep}) is adopted to solve these problems. 

As far as we know, different colors in the picture are represented by different numbers in the computer, so that we can treat the picture as a matrix of numbers. And the water supply data is an one-dimensional matrix composed of a series of numbers. Similarly, the water supply data can be considered as a special image. Inspired by transfer learning, the ResNet can be instinctively applied to the prediction of water supply, as shown in Fig \ref{fig_attention}.

\begin{figure}[htbp]
\begin{center}
\includegraphics[width=8.7cm]{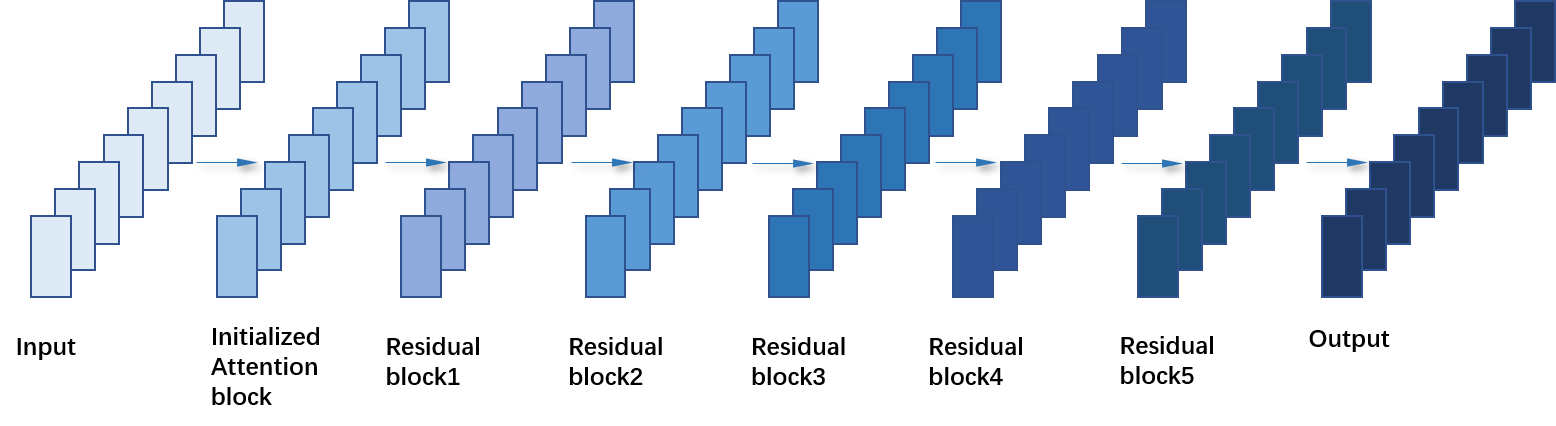}    
\caption{An overview of the IARN model: We use one attention block and five residual modules to stack our model. (Best viewed in color)} \label{fig_network}
\label{fig:bifurcation}
\end{center}
\end{figure}

As illustrated in Fig \ref{fig_network}, we employ the initialized attention module to obtain context over local features. Residual blocks are adopted as the backbone in our proposed network. The input and output are constrained to the same size between every block.
Firstly, we feed the input into the initialized attention module and generate new features of long-range contextual information on the following steps. 
The first step is to generate an initialized attention matrix. Next, we perform a matrix multiplication between the attention matrix and the original features. Finally, we perform an element-wise sum operation on the above multiplied resulting matrix and original features to obtain the final representations reflecting long-range contexts. 
Then we feed the new features into the Resnet model to get the senior semantic information. The best practice in convolutional network structure is distilled into a simplified architecture which can serve as a powerful point. In the following subsections, we will first introduce our new attention module and then explain the details and reasons for applying ResNet to the network.

\subsection{Initialized Attention}

In recent researches in the field of computer vision and natural language processing (NLP), attention module is one of the most popular network structure, such as `content-base attention' (\cite{Graves2014Neural}), `scaled dot-product attention' (\cite{brochier2019link}; \cite{vaswani2017attention}), `self-attention' (\cite{tan2018deep}; \cite{shen2018disan}; \cite{DBLP:journals/corr/abs-1802-10569}; \cite{DBLP:journals/corr/abs-1802-10569}) and `soft attention $\&$ hard attention' (\cite{xu2015show}). However, it is hard to find an attention model that is suitable for flow prediction tasks. In order to model complex contextual relationships of local features and reinforce the learning ability of the information, we introduce an initialized attention module, which can play a pivotal key as a preproceesing function.

In Computer Version field, image preprocessing is a significant part, whether in target monitoring, see \cite{ren2015faster}, \cite{DBLP:journals/corr/abs-1906-11172} and \cite{zhong2017random} or semantic segmentation, see \cite{ronneberger2015u}. Similarly, in time-series data prediction task, data preprocessing is also crucial. In this paper, we design an attention mechanism to enable intensive learning of input data, which is suitable for time series flow data prediction. To play the role of data preprocessing and encode a wider range of contextual information into local feature, the initialized attention module is employed on the top of the network. Important features and information can be enhanced while redundant information will be suppressing through this attention map. Next, we elaborate the process to adaptively aggregate the whole contexts.

\begin{figure}[htbp]
\begin{center}
\includegraphics[width=8.5cm]{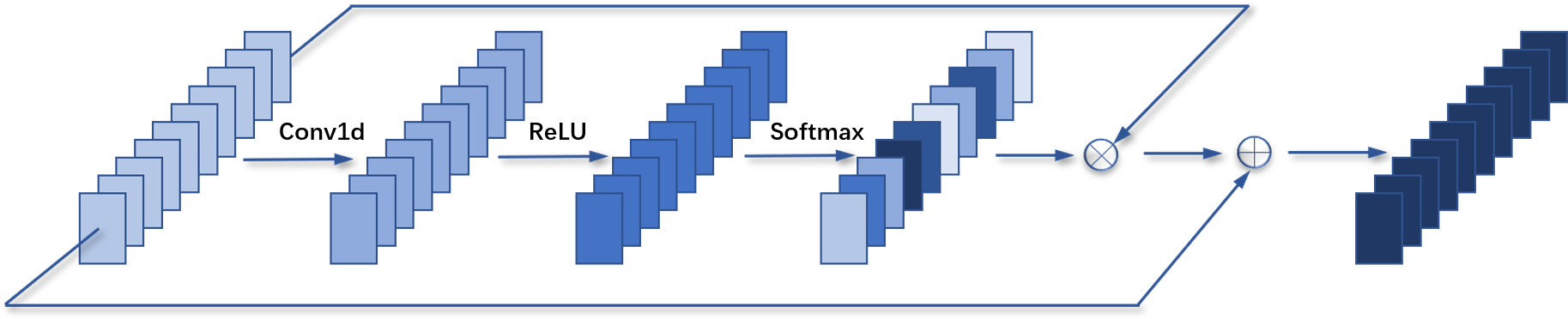}    
\caption{Illustration of attention block, $\oplus$ and $\otimes$ denote addition and multiplication layers, respectively. (Best viewed in color)} \label{fig_attention}
\label{fig:bifurcation}
\end{center}
\end{figure}

As shown in Fig \ref{fig_attention}, given an input A $\in \mathbb{R}^{C \times L}$, we first feed it into a convolution layers to generate a new feature map B, where B $\in \mathbb{R}^{C \times L}$. And then we employ a softmax layer to calculate the attention map S $\in \mathbb{R}^{C \times L}$:
\begin{equation}
S_{i}=\frac{e^{B_{i}}}{\sum_{i=1}^{n} e^{B_{i}}}
\end{equation}
Meanwhile, we multiply the attention map by A and use an element-wise sum operation with it to get the final output E $\in \mathbb{R}^{C \times L}$:
 \begin{equation}
E_{j}=\sum_{i=1}^{N}\left(s_{i} A_{i}\right)+A_{i}
\end{equation}\label{eq2}
Finally, we get the following formula (\ref{attention}):

\begin{equation}\label{attention}
E=\sum \frac{\exp (\max (0, y))}{\sum \exp (\max (0, y))} \otimes A \oplus A
\end{equation}
where $y$ is the result of the convolution of input, $\otimes$ is the multiplication of the corresponding elements and $\oplus$ means an element-wise sum. It can be seen that we repeatedly used the initial input in (\ref{attention}), and get our attention map through ReLU, convolution and softmax operations, which can strengthen the learning of input in the network and improve the data preprocessing ability. We thereby enhancing the network to capture global information and optimizing it by adding the attention layer. From now on, the model has an optimized global contextual view. 

\subsection{Residual Block}
In order to learn about the senior semantic information and capture more global information, deeper layers are required in our network. However, deeper networks are more likely to cause problems such as vanishing/exploding gradients and degradation problem, see \cite{BENGIO1994Learning}, \cite{He2015Convolutional} and \cite{srivastava2015highway}. So the residual blocks are considered in our network. In this paper, we define a residual block as:
\begin{equation}
\mathbf{y}=\mathcal{F}\left(\mathbf{x},\left\{W_{i}\right\}\right)+W_{s} \mathbf{x}
\end{equation}
where $W_{s}$ is used to match the dimensions, and $\mathcal{F}\left(\mathrm{x},\left\{W_{i}\right\}\right)$ denote the residual mapping to be studied. And the original mapping is recast into $\mathcal{F}(\mathrm{x})+\mathrm{x}$. This formulation can be realized by feedforward network with shortcut connections which are those skipping one or more layers, see \cite{Bishop:1995:NNP:525960}, \cite{bartkowiak2004neural} and \cite{schraudolph1998centering}). In our case, due to the shortcut only performs identity mapping, our model do not add extra parameters or computational complexity. And the entire network can still be trained end-to-end by optimizer with backpropagation.

\begin{figure}[htbp]
\begin{center}
\includegraphics[width=7.5cm]{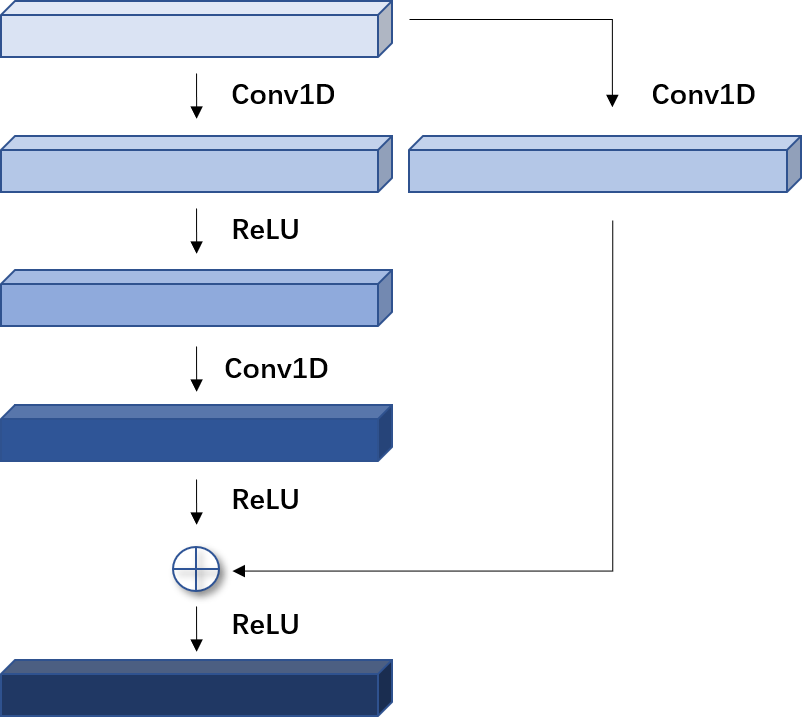}    
\caption{Illustration of the residual block. $\oplus$ represents element-wise sum layer in the network. (Best viewed in color)} \label{fig_res}
\label{fig:bifurcation}
\end{center}
\end{figure}

In order to obtain a network which produces an output of the same length as the input, we use a 1D fully-convolutional network (FCN) architecture, see \cite{long2015fully}. Since receptive field of our model depends on the network depth $n$ as well as filter size $k$, the stabilization of deeper model becomes extremely important. In our design of the generic model, we employ five residual blocks (Fig \ref{fig_res}), each of them has two convolutional layers and a rectified linear unit (ReLU) (\cite{nair2010rectified}):
\begin{equation}
o=\text { Activation }(\mathbf{x}+\mathcal{F}(\mathbf{x}))
\end{equation}
In standard ResNet, the input is added directly to the output of the residual function. However, there are different length of input and output. As a consequence, an extra $1 \times 1$ convolution is adopted to make sure element-wise addition $\oplus$ receives tensors of the same shape.

We now summarize the main characteristics of our network IARN as follow:
\begin{itemize}
    \item IARN handles time series tasks through the CNN structure. It utilizes the weight-sharing feature of convolutional layer and the interconnection between layers, which enables the model to be lightweight and improves the ability to learn data.
    \item IARN can boost the studying of crucial information and restrain the insignificance information by employing attention module at the beginning of the network. 
    \item IRAN enables the network to increase the depth of the network and learn senior semantic information in data without adding additional parameters. Problems such as gradient disappearance and gradient explosion are also avoided by adopting ResNet.
\end{itemize}

\section{Experiments}\label{section3}

To demonstrate the performance of IARN, we have tested them on three different water supply datasets, WSSS, WSCS and WSRC, collected by Shanghai Water Company. Each dataset contains key attributes of water flow with corresponding timestamps, as detailed below. We further show generalization of our approach and provide reliable results of our model.
\subsection{Datasets}

$\textbf{WSSS}$ was the water supply dataset which gathered from one area of southwest district of Shanghai. The water supply data are collected every day. The time period used from $4^{\text {th }}$ September, 2016 to $4^{\text {th }}$ May, 2019. We select the first two years of historical flow records as training set, and the rest serves as test set. This will ensure that all kinds of water flow data for the whole year can be learned by the model.

$\textbf{WSCS}$ was collected in real-time by several senior stations, deployed across the centre district of Shanghai. The dataset is aggregated into every day interval from every hour data samples. We select from $1^{\text {st }}$ January, $2017$ to $1^{\text {st }}$ September, $2019$. For this study, $80\%$ of the flow data are used for training and the remaining $20\%$ are used for testing. 

$\textbf{WSRS}$ collected the water supply data from a reservoir in Shanghai. The dataset is aggregated into every hour interval from every second data samples. The time range of WSRS dataset is $23^{th}$ October to $24^{th}$ September, $2019$. We split the training and test sets based on the same principles as above.
We do not use coarse data in our experiments.
\subsection{Implementation and evaluation}
Our proposed framework was implemented on Keras. Mean square error (MSE) was used for each output of this network. The initial learning rate was set to 0.001, and our deep models were optimized using the Adam optimizer with a momentum of 0.9 and a weight decay of 0.0005. Training time was set to 100 epochs, and batch size was set to 128 for all datasets. Then we tested our model among these three datasets we mentioned above. To completely evaluate our model and other RNN-based model, the following four methods were adopted:

Root mean square error (RMSE):
\begin{equation}\label{RMSE}
R M S E\left(y, y_{p r e}\right)=\sqrt{\frac{1}{n} \sum_{i=1}^{n}\left(y-y_{p r e}\right)^{2}}
\end{equation}
Mean absolute error (MAE):
\begin{equation}\label{MAE}
\operatorname{MAE}\left(y, y_{p r e}\right)=\frac{1}{n} \sum_{i=1}^{n}\left|y-y_{p r e}\right|
\end{equation}
Mean absolute percentage error (MAPE):
\begin{equation}\label{MAPE}
\operatorname{MAPE}\left(y, y_{p r e}\right)=\sum_{i=1}^{n}\left|\frac{y-y_{p r e}}{y}\right|\cdot \frac{100}{n}
\end{equation}
Explained variance score (EVS):
\begin{equation}\label{EVS}
\operatorname{EVS}\left(y, y_{p r e}\right)=1-\frac{\operatorname{Var}\left\{y-y_{p r e}\right\}}{\operatorname{Var}\{y\}}
\end{equation}
where $n$ is the number of the prediction data, $y$ is the actual value  and $y_{p r e}$ is the predicted value. To normalize the evaluation, (\ref{RMSE}) and (\ref{MAE}) are divided by $\frac{1}{n} \sum_{i=1}^{n} y$. Moreover, during our experiment, it can be found that generalization ability is positively correlated with EVS.


\subsection{Results On Dataset}

\begin{table*}[htbp]
\caption{Testing results of different networks on three datasets.}\label{tb:all}
\resizebox{\textwidth}{16mm}{
\begin{tabular}{ccccccccccccc}
\toprule
\multirow{2}{*}{\textbf{Methods}} & \multicolumn{4}{c}{\textbf{WSSS}} & \multicolumn{4}{c}{\textbf{WSCS}} & \multicolumn{4}{c}{\textbf{WSRS}}\\
\cmidrule(r){2-5}\cmidrule(r){6-9}\cmidrule(r){10-13}
& RMSE & MAE & MAPE($\%$) & EVS & RMSE & MAE & MAPE($\%$) & EVS & RMSE & MAE & MAPE($\%$) & EVS \\ \midrule
LSTM             & 0.116          & 0.098          & 5.15        & 0.364          & 0.031         & 0.026          & 1.34            & 0.496          & 0.191          & 0.157          & 9.94         & 0.443          \\
GRU              & 0.130         & 0.109          & 5.73            & 0.207          & 0.030          & 0.025          & 1.28            & 0.551          & 0.259          & 0.209          & 13.23            & 0.062          \\
Bi-LSTM          & 0.124          & 0.104          & 5.49            & 0.277          & 0.031          & 0.026          & 1.34            & 0.523          & 0.246          & 0.198          & 12.56            & 0.048          \\
Bi-GRU           & 0.155          & 0.130          & 6.84            & 0.133          & 0.044          & 0.038          & 1.94            & 0.036          & 0.255          & 0.205          & 13.01            & 0.025          \\
CNN-Bi-LSTM      & 0.139          & 0.119          & 6.25            & 0.104          & 0.042          & 0.036          & 1.87            & 0.094          & 0.248          & 0.199          & 12.63            & 0.036          \\
\textbf{IARN}    & \textbf{0.103} & \textbf{0.064} & \textbf{2.12} & \textbf{0.548} & \textbf{0.018} & \textbf{0.023} & \textbf{0.92} & \textbf{0.762} & \textbf{0.131} & \textbf{0.096} & \textbf{6.08} & \textbf{0.718} \\ \bottomrule
\end{tabular}}
\end{table*}

We compare our model with canonical recurrent architectures, namely: LSTM, GRU, Bi-LSTM, Bi-GRU and CNN-Bi-LSTM. These methods are first trained on 700 data records from these three dataset and the results have been shown in Table \ref{tb:all}. The comparative study shows that our network performs better than others in all evaluation metrics. On the WSSS dataset, our method achieves state-of-the-art performance in terms of MAE and RMSE, improving by $3.4\%$ and $1.3\%$ over strong RNN-based model. On the WSCS dataset, our method performs best with $2.3\%$ MAE and $1.8\%$ RMSE, when compared with recurrent neural network. As for WSRS dataset, which is totally different from the above two datasets, our network still exceeds RNN-based network by $6.1\%$ MAE and $6.0\%$ RMSE.

In practice, the water plant needs to carry out real-time update data and online prediction, so the recent data will be used for training and correction. In the small sample size of the dataset, generic recurrent models can not effectively deal with these data due to several factors such as the loss of segmental information in the forgetting gate, so the results are all performing poorer than IARN. Meanwhile, the forecast results are impacted by the influence of weather or other factors. As a result, it is difficult to obtain very accurate prediction result by using RNN-based models. However, our proposed network can learn the senior semantic information through the convolution structure and retain all the information which we input, so it is easier to get extremely preeminent results.

\begin{table}[htbp]
\begin{center}
\caption{Testing results on WSSS-a dataset. (WSSS dataset with abnormal data)}\label{tb:wsms}
\begin{tabular}{ccccc}
\toprule
\textbf{Methods} & \textbf{RMSE}   & \textbf{MAE}    & \textbf{MAPE($\%$)}   & \textbf{EVS}    \\ \midrule
LSTM             & 0.190          & 0.160          & 5.39          & 0.433          \\
GRU              & 0.175          & 0.147          & 4.96          & 0.519          \\
Bi-LSTM          & 0.183          & 0.154          & 5.20          & 0.488          \\
Bi-GRU           & 0.239          & 0.203          & 6.83          & 0.118          \\
CNN-Bi-LSTM      & 0.244          & 0.205          & 6.91          & 0.041          \\
\textbf{IARN}    & \textbf{0.103} & \textbf{0.069} & \textbf{2.36} & \textbf{0.813} \\ \bottomrule
\end{tabular}
\end{center}
\end{table}

\begin{table}[htbp]
\caption{Testing results on WSCS using WSSS-a-trained models.}
\label{tb:04}
\begin{tabular}{ccccc}
\toprule
\textbf{Methodsl} & \textbf{RMSE}   & \textbf{MAE}    & \textbf{MAPE($\%$)}   & \textbf{EVS}    \\ \midrule
LSTM           & 0.157          & 0.124          & 6.27          & 0.426          \\
GRU            & 0.145          & 0.147          & 4.96          & 0.519          \\
Bi-LSTM        & 0.183          & 0.488          & 15.43          & 0.052          \\
Bi-GRU         & 0.195          & 0.153          & 7.74          & 0.114          \\
CNN-Bi-LSTM    & 0.192          & 0.150          & 7.60          & 0.041          \\
\textbf{IARN}           & \textbf{0.025} & \textbf{0.018} & \textbf{0.18} & \textbf{0.829} \\ \bottomrule
\end{tabular}
\end{table}

\begin{table}[htbp]
\caption{Testing results on WSCS using WSSS-a-trained models.}
\label{tb:05}
\begin{tabular}{ccccc}
\toprule
\textbf{Methods} & \textbf{RMSE}   & \textbf{MAE}    & \textbf{MAPE($\%$)}   & \textbf{EVS}    \\ \midrule
LSTM           & 0.379          & 0.323          & 20.50          & 0.418          \\
GRU            & 0.380          & 0.324          & 20.47          & 0.415          \\
Bi-LSTM        & 0.361          & 0.310          & 19.59          & 0.471          \\
Bi-GRU         & 0.431          & 0.368          & 23.27          & 0.245          \\
CNN-Bi-LSTM    & 0.488          & 0.410          & 25.97          & 0.037          \\
\textbf{IARN}           & \textbf{0.131} & \textbf{0.099} & \textbf{1.25} & \textbf{0.730} \\ \bottomrule
\end{tabular}
\end{table}

\begin{figure}[htbp]
\begin{center}
\includegraphics[width=9cm]{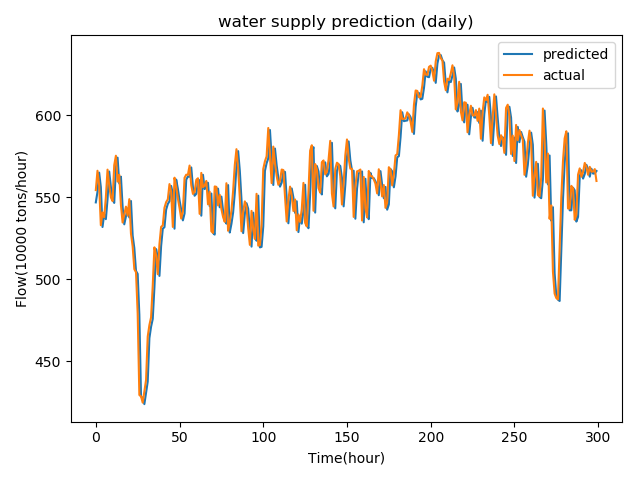}    
\caption{Generalization test on WSCS dataset. 300 continuous data are applied as test set.} \label{fig_WSCS}
\label{fig:bifurcation}
\end{center}
\end{figure}

\begin{figure}[htbp]
\begin{center}
\includegraphics[width=9cm]{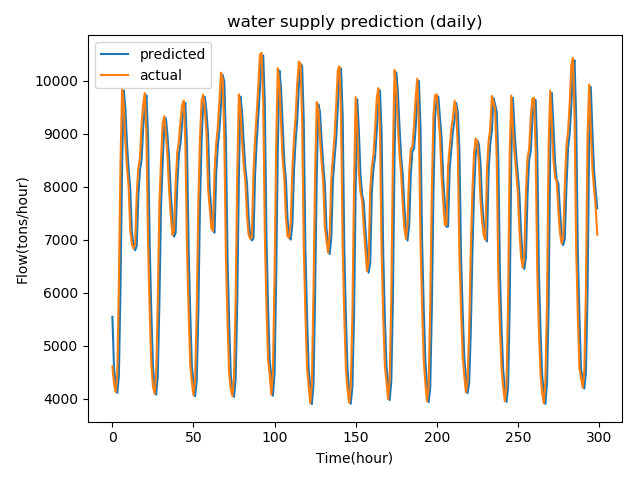}    
\caption{Generalization test on WSRS dataset. 300 continuous data are used as test set.} \label{fig_WSJS}
\label{fig:bifurcation}
\end{center}
\end{figure}

Ultimately, we conducts several studies on WSSS-a dataset, which is composed of normal data and abnormal data (abnormal data are caused by typhoon and other factors). In these experiments, methods are trained on WSSS-a dataset, and tested on WSCS and WSRS datasets. Our focus is on the behaviors of generalization ability but not on pushing the state-of-the-art results. As we can see from Table \ref{tb:wsms}, our model is particularly significant for the data with abnormal information, while our model preforms an outstanding generalization ability in Table \ref{tb:04} and Table \ref{tb:05}.  For further verification, results are then visualized (Fig \ref{fig_WSCS} and Fig \ref{fig_WSJS}). The curve of predicted value and true value are very close in two different datasets.


\begin{figure}[htbp]
\begin{center}
\includegraphics[width=9cm]{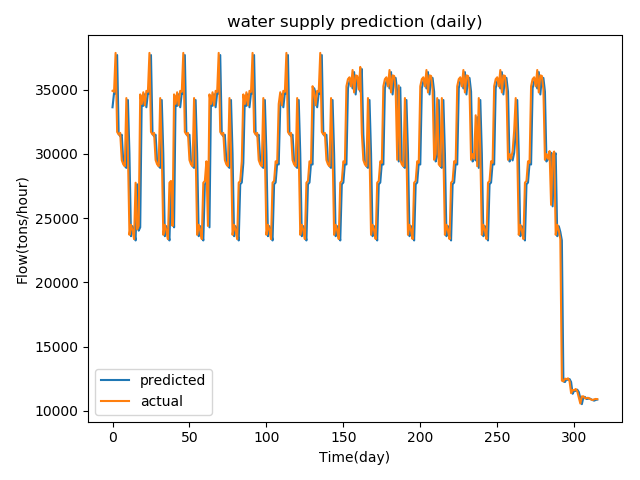}    
\caption{Result on WSSS dataset with abnormal data. The abnormal data can be found at the end of the curve.} \label{fig_WSMS}
\label{fig:bifurcation}
\end{center}
\end{figure}

\section{Conclusion}\label{section4}

In this paper, we have presented a novel convolutional network IARN for water supply prediction. Attention module and residual module are adopted in our proposed model. Experiments have shown that our framework outperforms other single methods and achieves the state-of-the-art performance on three different datasets. It also performs better robustness and generalization ability. These features are quite promising and practical for scholarly development and large-scale industry deployment. In the future, we will optimize the structure of the proposed network and apply our model into other flow forecasting scenarios, such as Web traffic and electricity consumption.

\bibliography{ifacconf}             
                                                   







\end{document}